\documentclass[letterpaper, 10 pt, conference]{ieeeconf}  
\usepackage{graphicx}
\usepackage{amsmath}
\usepackage{amssymb}
\usepackage{multirow}
\usepackage{booktabs}
\usepackage{tabularx}
\usepackage{array}
\usepackage{cuted}
\preCutedStrip{\vspace{-1cm}}

\IEEEoverridecommandlockouts                              

\title{\LARGE \bf
ExcavaTwin: Training-Free Geometry-Guided Semantic Elevation Mapping for Autonomous Excavation
}

\author{Yu Deng$^{1}$, Lingshan Zeng$^{2}$, Tong Hu$^{3}$, Rushi Dai$^{1,*}$ 
\thanks{*Corresponding author.}
\thanks{This work was supported in part by the Guangdong Provincial Key Laboratory of Integrated Communication, Sensing and Computation for Ubiquitous Internet of Things (No. 2023B1212010007).}%
\thanks{$^{1}$ The Smart Manufacturing Thrust,
The Hong Kong University of Science and Technology (Guangzhou)
        {\tt\small }}%
\thanks{$^{2}$ Department of Electronic and Electrical Engineering,
Southern University of Science and Technology.
        {\tt\small}}%
        \thanks{$^{3}$ Capstone Technology Co., Ltd., Shenzhen, China.
        {\tt\small}}%
}

\begin{document}

\maketitle
\thispagestyle{empty}
\pagestyle{empty}

\begin{strip}
    \centering
    \includegraphics[
        width=0.90\textwidth
    ]{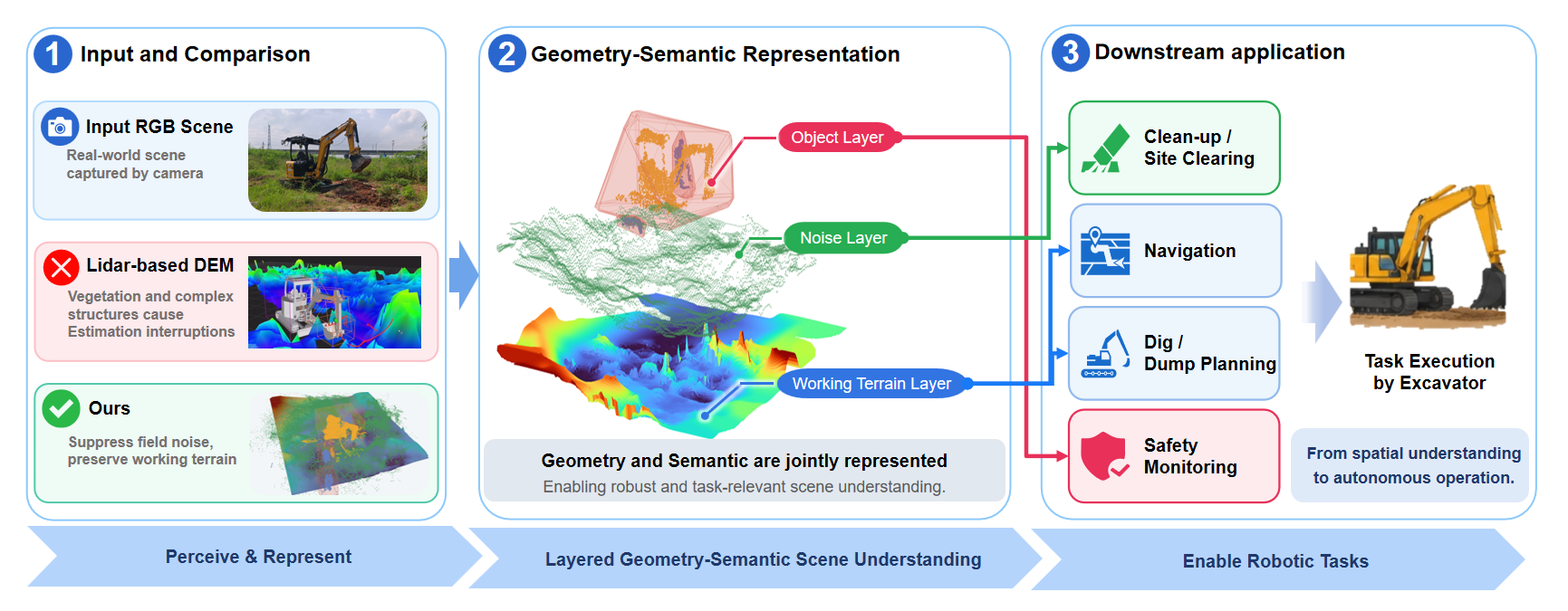}

    \vspace{0.2cm}

    \refstepcounter{figure}
    {\small Fig.~\thefigure. Overview of ExcavaTwin. LiDAR-based perception in outdoor environments can be affected by environmental interference. ExcavaTwin uses visual input to construct a clear, multi-layer spatial representation for excavators' tasks.
}
    \label{fig:overview}
\end{strip}


\begin{abstract}
Autonomous excavation requires a spatial representation that jointly captures terrain geometry and task-relevant semantics. Existing excavation mapping is largely elevation-centric, while generic semantic models remain unstable in unstructured outdoor scenes. We present ExcavaTwin, a pure-vision geometry-guided semantic elevation mapping framework without excavation-specific training. Given multi-view RGB images, the framework: 1) reconstructs scene geometry and semantic observations using frozen vision models; 2) derives terrain and non-terrain geometric support; 3) performs geometry-constrained multi-view semantic fusion to suppress implausible predictions and recover incomplete observations; and 4) projects the fused state into a task-oriented semantic elevation map. Experiments on public datasets and real excavation scenes demonstrate reliable geometric and semantic perception. In real excavation, the system achieved an average update interval of approximately 1.4 s and a mean elevation error of 12.74cm in dynamically modified regions. Larger errors mainly occur during rapid terrain changes and transient visual disturbances caused by machine motion.
\end{abstract}

\section{Introduction}

Autonomous excavation requires a spatial representation that identifies where
the machine can operate, what parts of the scene should be avoided, and how the
terrain changes during execution. Conventional elevation maps provide compact
estimates of surface height and uncertainty~\cite{fankhauser2014elevation,miki2022elevation,jud2019trenching,jud2021heap},
but do not distinguish operable terrain from surface interference, machinery,
or obstacles. Consequently, non-terrain observations may be incorporated into
the estimated surface, while geometrically adjacent regions with different
task roles remain indistinguishable.

Projecting image semantics into 3D does not fully resolve this limitation.
Single-view predictions are incomplete under occlusion and may vary across
viewpoints, while generic semantic categories do not necessarily reflect the
functional roles required for excavation. Direct multi-view voting improves
consistency by accumulating semantic scores at corresponding 3D
locations~\cite{mccormac2017semanticfusion,yamazaki2024openfusion}, but
it does not determine whether a predicted class is physically compatible with
the local terrain or object structure. As a result, repeated but geometrically
implausible predictions may still be retained, and uncertain regions may be
forced into an observed class.

We present ExcavaTwin, a pure-vision geometry-guided semantic elevation mapping
framework that requires no excavation-specific semantic training.  Unlike direct
3D voting, which selects labels primarily from accumulated per-view scores, our
method restricts the candidate label space using connected terrain/non-terrain
support and class-dependent geometric admissibility, while retaining an
explicit unknown state when the available evidence is insufficient or
conflicting. Geometry and semantics therefore contribute complementary
constraints to a shared scene interpretation. An overview of the proposed
framework is shown in Fig.~\ref{fig:overview}.

Given multi-view RGB observations, the proposed framework first reconstructs
scene geometry and obtains semantic observations using frozen vision models.
It then extracts connected terrain and non-terrain support from the
reconstruction. Semantic observations are subsequently aligned and fused
across views under class-dependent geometric constraints. Finally, the fused
state is mapped to an excavation-oriented taxonomy and projected into a metric
semantic elevation map containing terrain geometry, semantic identity, and
observation confidence.

The main contributions of this work are:
\begin{itemize}
    \item A vision-based geometry--semantic elevation mapping framework for autonomous excavation that requires no specific semantic training or fine-tuning.

    \item A geometry-guided fusion mechanism that improves semantic segmentation performance in outdoor environments without requiring additional training.

    \item Extensive evaluation across public datasets, the OCES dataset, and real-world excavation, demonstrating geometric accuracy, semantic robustness, and practical feasibility.
\end{itemize}

\begin{figure*}[t]
    \centering
    \includegraphics[width=18cm,
    trim={0 0cm 0 0cm},
    clip]{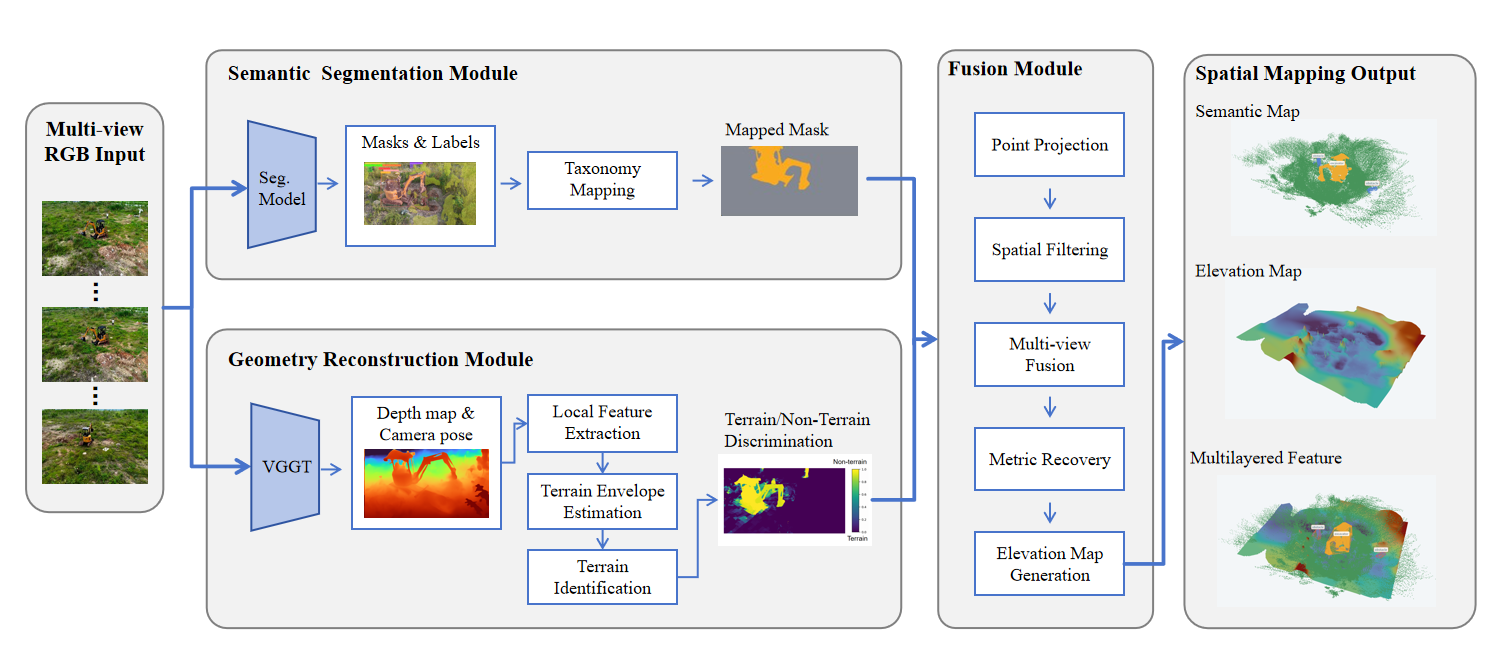}
    \caption{Main pipeline of ExcavaTwin. The pipeline consists of two stages: vision-based twinning and geometry--semantic coupling. First, frozen visual reconstruction and segmentation models, such as VGGT and YOLOE, recover the scene structure and semantics. Then, geometric features are fused with semantic information to produce an excavation-oriented spatial map with semantic and elevation representations.
}
    \label{fig:pipeline}
    \vspace{-0.6cm} 
\end{figure*}

\section{Related Work}
\label{sec:related_work}

\subsection{Terrain Mapping for Field Robotics}

Terrain mapping has been extensively studied for field robotics, where 2.5D elevation representations provide an efficient description of local surface geometry. Fankhauser et al.~\cite{fankhauser2014elevation} introduced probabilistic robot-centric elevation mapping with explicit uncertainty modelling, and subsequent work improved its efficiency and applicability to rough-terrain locomotion~\cite{miki2022elevation}. For excavation, Jud et al.~\cite{jud2019trenching,jud2021heap} extended terrain mapping to account for self-occlusion and continuously modified surfaces during digging, while Bhandari et al.~\cite{bhandari2025dynamic} further introduced probabilistic temporal modelling for rapidly changing terrain. These studies demonstrate that geometric terrain estimation and continuous updating are already well established. However, the maintained scene state remains dominated by geometric quantities such as elevation, local shape, and uncertainty, while semantic identity is typically handled separately.

\subsection{Outdoor Semantic Perception and Open-Vocabulary Models}

Semantic perception in unstructured outdoor environments remains more dependent on domain-specific data. Dedicated datasets such as RELLIS-3D~\cite{jiang2020rellis3d} and WildScenes~\cite{vidanapathirana2025wildscenes} were introduced specifically to support semantic understanding in off-road and natural environments, reflecting the difficulty of directly transferring models trained on structured urban scenes. For construction and excavation sites, this dependency is particularly restrictive because soil appearance, vegetation, illumination, machine occlusion, and terrain deformation vary substantially across worksites. Recent open-vocabulary models alleviate this requirement by enabling zero-shot recognition of previously unseen concepts. YOLOE~\cite{wang2025yoloe}, for example, provides prompt-based open-set detection and segmentation without excavation-specific training. Nevertheless, such predictions remain image-space observations and may be incomplete or inconsistent across viewpoints. Thus, eliminating task-specific training does not by itself produce a persistent and physically coherent scene representation.

\subsection{3D Semantic Mapping and Geometry--Semantic Fusion}

A related line of research integrates image semantics with reconstructed 3D geometry. SemanticFusion~\cite{mccormac2017semanticfusion} demonstrated multi-view fusion of CNN predictions through dense geometric correspondences, improving semantic consistency over single-frame results. More recent methods further extend semantic mapping toward open-vocabulary scene understanding. OpenScene~\cite{peng2023openscene} transfers vision-language features into 3D for zero-shot semantic queries, while ConceptFusion~\cite{jatavallabhula2023conceptfusion} and Open-Fusion~\cite{yamazaki2024openfusion} construct persistent 3D representations from pretrained foundation-model features without task-specific fine-tuning. These methods significantly reduce the dependence on fixed semantic taxonomies and annotated 3D training data.

However, existing semantic mapping methods mainly use geometry as a spatial carrier for accumulating semantic predictions. In unstructured earthmoving scenes, geometry can provide stronger physical constraints: terrain continuity, local relief, surface orientation, and 3D object support indicate where particular semantic hypotheses are physically plausible. Our method therefore uses reconstructed geometry not only to associate multi-view semantic observations, but also to constrain and complete incomplete open-vocabulary predictions over the observed physical surface. The resulting geometry--semantic scene state is further projected into a task-oriented elevation representation for autonomous excavation, where semantic identity, terrain structure, and observation validity are jointly maintained. This differs from semantic elevation mapping approaches that rely on site-specific semantic training~\cite{chen2026semantic}, while retaining the compact terrain representation required by downstream excavation tasks.


Learning-based reconstruction models provide dense depth, camera geometry, and
three-dimensional point predictions from a small set of
images~\cite{vggt}.  Their output,
however, remains an unstructured geometric observation: it does not state which
points form a continuous terrain surface, which structures are physically
detached from that surface, or how semantic evidence should be interpreted at
terrain--object boundaries.  Conversely, an image segmenter can recognize
appearance, but its predictions are view dependent, incomplete under
occlusion, and unaware of the physical support of a class.  Our objective is
therefore not to attach independent two-dimensional labels to a reconstructed
point cloud.  Instead, we seek a physically supported scene decomposition in
which terrain structure and semantic identity are inferred on the same
three-dimensional support.

\section{Method}
\subsection{Problem Formulation}
\label{sec:gc_problem}

As illustrated in Fig.~\ref{fig:pipeline}, the framework takes a sequence of RGB observations
and jointly builds geometric and semantic representations. Given a sequence of RGB observations
$\mathcal I=\{I_s\}_{s=1}^{S}$, a geometry backbone such as VGGT
predicts a world point map $P_s$, camera parameters, and a reconstruction
confidence map $C_s$ for each view~\cite{vggt}.  In parallel, an
open-vocabulary segmenter~\cite{wang2025yoloe} provides a non-negative semantic evidence vector
$\mathbf o_s(\mathbf u)$ at image location $\mathbf u$.
The point prediction and semantic observation are retained on the same image
lattice.  Each valid image sample therefore defines
\begin{equation}
  (s,\mathbf u)
  \longleftrightarrow
  \bigl(\mathbf p_{s\mathbf u},c_{s\mathbf u},
  \mathbf o_{s\mathbf u}\bigr),
  \label{eq:gc_correspondence}
\end{equation}
where $\mathbf p_{s\mathbf u}$ is the reconstructed point and
$c_{s\mathbf u}$ is its reconstruction confidence.  This is an index-level
correspondence, not a separately estimated nearest-neighbor association
between a two-dimensional mask and an unrelated point cloud.

The framework integrates these geometric and semantic
observations and projects the fused representation onto
a horizontal grid to construct a semantic elevation map:
\begin{equation}
\mathcal{M}
=
\mathcal{F}
\left(
\left\{
(\mathbf{p}_{s\mathbf{u}},
c_{s\mathbf{u}},
\mathbf{o}_{s\mathbf{u}})
\right\}_{s,\mathbf{u}}
\right),
\label{eq:semantic_elevation_map}
\end{equation}
where $\mathcal{F}$ denotes the proposed geometry--semantic
mapping framework and $\mathcal{M}$ is the resulting semantic
elevation map.

\subsection{Task-Oriented Hierarchical Spatial Taxonomy}
\label{sec:gc_taxonomy}

Autonomous excavation requires distinguishing manipulable terrain,
environmental interference, and objects relevant to operational
safety. We therefore organize the reconstructed scene into
task-oriented functional classes rather than relying directly
on the predefined vocabulary of generic segmentation models.

We partition the workspace into three task-oriented layers:
an object layer for machinery and obstacles, a noise layer for
environmental interference, and a working terrain layer for
manipulable ground.
The proposed framework uses geometric structure to constrain semantic inference, suppressing implausible predictions while semantic evidence resolves the identities of spatially admissible regions.The resulting geometry--semantic representation is projected
onto a horizontal grid to construct a multi-layer semantic
elevation map,
\begin{equation}
\mathcal{M}
=
\left\{
\mathcal{M}^{\mathrm{work}},
\mathcal{M}^{\mathrm{noise}},
\mathcal{M}^{\mathrm{object}}
\right\},
\label{eq:multilayer_map}
\end{equation}
where the three layers respectively preserve terrain elevation,
environmental interference, and task-relevant objects within
a common spatial reference.

Geometric structure first establishes terrain and non-terrain
support, constraining the spatial admissibility of semantic
predictions. Within these supports, semantic evidence further
distinguishes construction terrain, other terrain, removable
surface cover, excavators, and obstacles. Construction terrain
represents manipulable soil, other terrain provides surrounding
spatial reference, and removable surface cover represents
environmental interference such as vegetation. Excavators and
obstacles represent machinery and objects relevant to operational
safety, respectively. Regions with insufficient or conflicting
evidence are assigned to the unknown class. Generic semantic
predictions are mapped to these task-oriented classes to provide
a consistent interpretation of the reconstructed scene.

Following this taxonomy, the outputs of the semantic segmentation
model are mapped to the corresponding task-oriented classes using
a predefined class mapping. Predictions assigned to the same task
class are merged into a common semantic evidence channel, providing
a unified label space for subsequent geometry-constrained fusion.

\subsection{Geometric Spatial Processing}
\label{sec:gc_geometry}
The geometric branch constructs the spatial support for subsequent semantic inference rather than determining semantic identity independently. Starting from the reconstructed point cloud, we first extract local geometric features and identify unreliable observations. These features are then used to establish a continuous terrain support, from which the remaining non-terrain regions are separated and organized into object-level components. The resulting terrain and object supports provide the geometric basis for subsequent geometry-constrained semantic fusion.

\subsubsection{ Local Geometric Feature Extraction}
\label{sec:gc_local_geometry}

From the three-dimensional point cloud produced by visual reconstruction, we
first extract local geometric structure to establish a stable geometric
representation for subsequent spatial reasoning. The raw point cloud is
voxelized at a scene-adaptive resolution, and a $k$-nearest-neighbor topology
is built over the voxel centers using a spatial index to define consistent
local neighborhoods $\mathcal{N}_i$. Following covariance-based surface
analysis~\cite{pauly2002simplification}, we perform an eigendecomposition of
the local covariance matrix within each neighborhood,
\begin{equation}
\mathbf{C}_i = \frac{1}{|\mathcal{N}_i|}
\sum_{j \in \mathcal{N}_i} (\mathbf{p}_j - \bar{\mathbf{p}}_i)
(\mathbf{p}_j - \bar{\mathbf{p}}_i)^{\top}
= \sum_{m=1}^{3} \lambda_{i,m}\, \mathbf{v}_{i,m} \mathbf{v}_{i,m}^{\top}
\label{eq:cov}
\end{equation}
where $\bar{\mathbf{p}}_i$ is the neighborhood centroid. The eigenvector
$\mathbf{v}_{i,1}$ associated with the smallest eigenvalue is taken as the
surface normal $\mathbf{n}_i$ (consistently oriented upward), while the
eigenvalue spectrum yields planarity descriptors such as the
orthogonal-residual roughness $r_i = \sqrt{\lambda_{i,1}}$ and the curvature
$\kappa_i = \lambda_{i,1} / (\lambda_{i,1} + \lambda_{i,2} + \lambda_{i,3})$.
From the normal we derive the local slope
$\theta_i = \arccos\!\big(\mathbf{n}_i \cdot \hat{\mathbf{z}}\big)$ as its
deviation from the vertical axis $\hat{\mathbf{z}}$, and from the height
distribution within each neighborhood we compute the raw height variance
together with a detrended elevation discontinuity, defined as the median
absolute distance to the local tangent plane, which isolates genuine surface
undulation from smooth slope.

Rather than committing points to premature outlier decisions, we identify
unreliable observations through spatial consistency. Let $d_i$ denote the mean
distance from voxel $i$ to its neighbors; using the median $\tilde{d}$ and the
median absolute deviation $\mathrm{MAD}(d)$ as a robust scale
estimate~\cite{hampel1974influence}, a point
is flagged as a spatial outlier when
\begin{equation}
d_i > \tilde{d} + \gamma \cdot \mathrm{MAD}(d),
\quad
\mathrm{MAD}(d)
=\operatorname{median}_i
\left| d_i - \tilde{d} \right|
\label{eq:outlier}
\end{equation}
with sensitivity parameter $\gamma$. This robust criterion excludes
geometrically incoherent points from all downstream geometric support.

\subsubsection{Terrain Envelope Estimation}
\label{sec:gc_terrain_support}

Based on the local geometric evidence, we partition the scene into continuous
terrain and non-terrain structures. Because legitimate terrain in an excavation
scene may be sloped or locally uneven, we do not assume a single global ground
plane. Instead, inspired by progressive morphological ground
filtering~\cite{zhang2003progressive}, we estimate a local terrain reference
surface over the horizontal plane, and form an initial terrain estimation from
each point's height relative to this surface, together with its local slope and
its spatial continuity. 
Pointwise compatibility alone is not sufficient, since isolated
elevated fragments can also satisfy these conditions; we therefore impose a neighborhood elevation-continuity constraint. Two adjacent cells $a$ and $b$ are linked into the same terrain region only when their reference elevations vary continuously,
\begin{equation}
  \big| \widehat H(a) - \widehat H(b) \big|
  \leq \delta_h L + \delta_s\, d(a,b),
  \label{eq:gc_continuity}
\end{equation}
where $d(a,b)$ is their horizontal spacing and the slope term $\delta_s$
tolerates gradual elevation change. Isolated and small connected components are
then discarded. This yields a spatially continuous terrain region, while the
remaining structures are treated as non-terrain and passed to subsequent
analysis.

\subsubsection{Non-Terrain Object Extraction}
\label{sec:gc_object_support}

With the continuous terrain support identified, we restrict object extraction to the remaining non-terrain points. This separation reduces ground-induced connections during spatial clustering and allows distinct object structures to be recovered from the reconstructed scene. We then apply reliability filtering, spatial clustering, and geometric cleanup to obtain object-level geometric support.

For each class $c$, we retain points whose reconstruction confidence and class-specific evidence exceed their respective thresholds.
 A point $i$ of class
$c_i$ is retained when
\begin{equation}
  \phi_i \;\geq\; Q_{q}\!\big(\{\phi_j : c_j = c_i\}\big)
  \quad\text{and}\quad
  e_i(c_i) \;\geq\; e_{\min},
  \label{eq:gc_conf_filter}
\end{equation}
where $\phi_i$ is the reconstruction confidence, $Q_{q}(\cdot)$ is the $q$-th
percentile taken within the class, and $e_i(c_i)$ is the accumulated evidence
for the point's own class. 

The surviving points are grouped into connected components on a
scene-relative radius graph, so that distinct machines and obstacles fall into separate components.

\subsection{Geometry-Semantic Fusion }
\label{sec:gc_semantic_fusion}

Our geometry--semantic fusion consists of three stages:
pixel-to-point semantic projection, spatial consistency filtering,
and reliability-weighted multi-view fusion with geometric constraints.
The resulting semantic 3D representation is subsequently projected
onto a horizontal grid to construct a semantic elevation map.

For each image, semantic responses are expressed in the task-oriented
taxonomy defined above. The response $o_s(\mathbf{u},k)$ for class $k$
at pixel $\mathbf{u}$ in view $s$ is associated with the reconstructed
3D point observed at the same image location,
\begin{equation}
o_s(\mathbf{u},k) \rightarrow e_{i,s}(k),
\end{equation}
where $e_{i,s}(k)$ denotes the semantic evidence assigned to point $i$
from view $s$.

Per-view segmentation produces incomplete and noisy semantic
observations after 3D projection. We address these inconsistencies
in the shared 3D cloud. For each class, labeled points are grouped
by single-linkage clustering with a distance threshold $\varepsilon$,
and clusters below a minimum size are discarded as noise.
The surviving clusters provide foreground anchors for propagating
labels to nearby unlabeled points. A candidate point receives a
class only when it is closer to a foreground anchor of that class
than to any corresponding background anchor. This foreground--background
competition fills gaps in semantic coverage while limiting label
propagation onto adjacent terrain. The filtered evidence is then
used for multi-view fusion.

The observations are aggregated on a common 3D voxel grid, so that
observations falling within voxel $j$ share the same spatial support.
Evidence from the same view is first consolidated into a single
class-wise vector $\mathbf{e}_{j,s}$, preventing densely sampled
views from contributing disproportionately. Each view contributes
a reliability weight $w_{j,s}$ incorporating semantic confidence,
reconstruction confidence, viewing angle, and camera distance.
The fused semantic vote is
\begin{equation}
v_j(k) =
\frac{\sum_s w_{j,s} e_{j,s}(k)}
     {\sum_s w_{j,s}+\epsilon},
\label{eq:vote}
\end{equation}
where $\epsilon$ is a small constant for numerical stability.

The semantic vote is combined with a class-dependent geometric
support term $g_j(k)$, which suppresses hypotheses inconsistent
with the local spatial structure. The normalized class score is
\begin{equation}
q_j(k) =
\frac{
g_j(k)\bigl[v_j(k)+\epsilon\bigr]^{\lambda_v}
}{
\sum_{\ell}
g_j(\ell)\bigl[v_j(\ell)+\epsilon\bigr]^{\lambda_v}
},
\label{eq:score}
\end{equation}
where $\lambda_v$ controls the contribution of semantic evidence.
The final label is selected as
\begin{equation}
\hat{k}_j =
\begin{cases}
\displaystyle \arg\max_k q_j(k),
& \displaystyle \max_k q_j(k) \geq \tau,\\
\mathrm{unknown}, & \text{otherwise},
\end{cases}
\label{eq:semantic_decision}
\end{equation}
where $\tau$ is the decision threshold.

To construct the semantic elevation map, the labeled 3D points
are projected onto a horizontal grid in the gravity-aligned
coordinate frame. Let $\mathcal{P}_c$ denote the points assigned
to cell $c$, and let $\mathcal{T}_c \subseteq \mathcal{P}_c$
contain points classified as construction terrain or other terrain
that satisfy the geometric terrain-support criterion.
The terrain elevation is estimated as
\begin{equation}
H_c = \operatorname{median}
\{h_i \mid i \in \mathcal{T}_c\},
\label{eq:semantic_elevation}
\end{equation}
where $h_i$ is the gravity-aligned height of point $i$.
The cell-level semantic label is obtained by confidence-weighted
voting,
\begin{equation}
L_c =
\arg\max_{k \in \mathcal{K}}
\sum_{i \in \mathcal{P}_c}
\alpha_i\,\mathbb{I}(\hat{k}_i=k),
\qquad
\alpha_i=\max(\gamma_i,\gamma_{\min}),
\label{eq:grid_semantics}
\end{equation}
where $\mathcal{K}$ includes the task-oriented classes and unknown,
$\hat{k}_i$ and $\gamma_i$ denote the assigned semantic label and
confidence of point $i$, and $\gamma_{\min}$ is a minimum voting
weight. Excavator and obstacle points additionally define object
occupancy and vertical extents within the same grid. Cells without
terrain support retain an undefined terrain elevation, while
observation masks distinguish observed and unobserved regions.
These co-registered layers provide a compact 2.5D representation
of terrain geometry, semantic classes, and object occupancy for
subsequent excavation-oriented analysis.

Finally, metric scale is recovered using a selected reference object
with a known physical dimension. Let $d_{\mathrm{ref}}$ denote this
dimension and $\hat{d}_{\mathrm{ref}}$ the corresponding measurement
in the reconstructed scene. A global scale factor is computed as
$s = d_{\mathrm{ref}}/\hat{d}_{\mathrm{ref}}$ and applied uniformly
to the reconstructed geometry. This converts the scene into metric
units while preserving its relative geometric structure, enabling
physical measurements of object dimensions and terrain elevation.

\section{Experiments}

\subsection{Experimental Details}
\label{sec:exp_setup}
\paragraph{Datasets}
We conducted experiments on four datasets:
RELLIS-3D~\cite{jiang2020rellis3d},
Objectron~\cite{ahmadyan2021objectron},
GOOSE-Ex~\cite{hagmanns2025gooseex}, and the collected OCES dataset.
These datasets cover outdoor terrain, object-level geometry, and
excavation-scene semantics, respectively, and are used to evaluate the
geometric and semantic capabilities of our framework. The characteristics
and evaluation roles of the four datasets are summarized in
Tab.~\ref{tab:dataset_comparison}.

\textbf{RELLIS-3D:}
RELLIS-3D is a large-scale multimodal off-road dataset containing synchronized RGB images and LiDAR measurements collected in unstructured outdoor environments. It provides annotations for 13,556 LiDAR scans and 6,235 RGB images across 20 semantic classes. We use selected scenes to evaluate outdoor geometric reconstruction and elevation estimation, with LiDAR measurements serving as the geometric reference~\cite{jiang2020rellis3d}.

\textbf{Objectron:}
Objectron is a large-scale object-centric dataset containing 14,819 annotated video sequences across nine common object categories. We sample 20 videos per category, resulting in 180 videos for evaluation~\cite{ahmadyan2021objectron}.

\textbf{GOOSE-Ex:}
GOOSE-Ex is an outdoor dataset designed for semantic perception in unstructured environments, providing pixel-level semantic annotations for diverse terrain and object classes. 
We evaluate our framework on the selected sequences of GOOSE-Ex to assess its semantic segmentation accuracy and its ability to suppress environmental interference~\cite{hagmanns2025gooseex}.

\begin{figure}[h]
    \centering
    \includegraphics[width=\columnwidth]{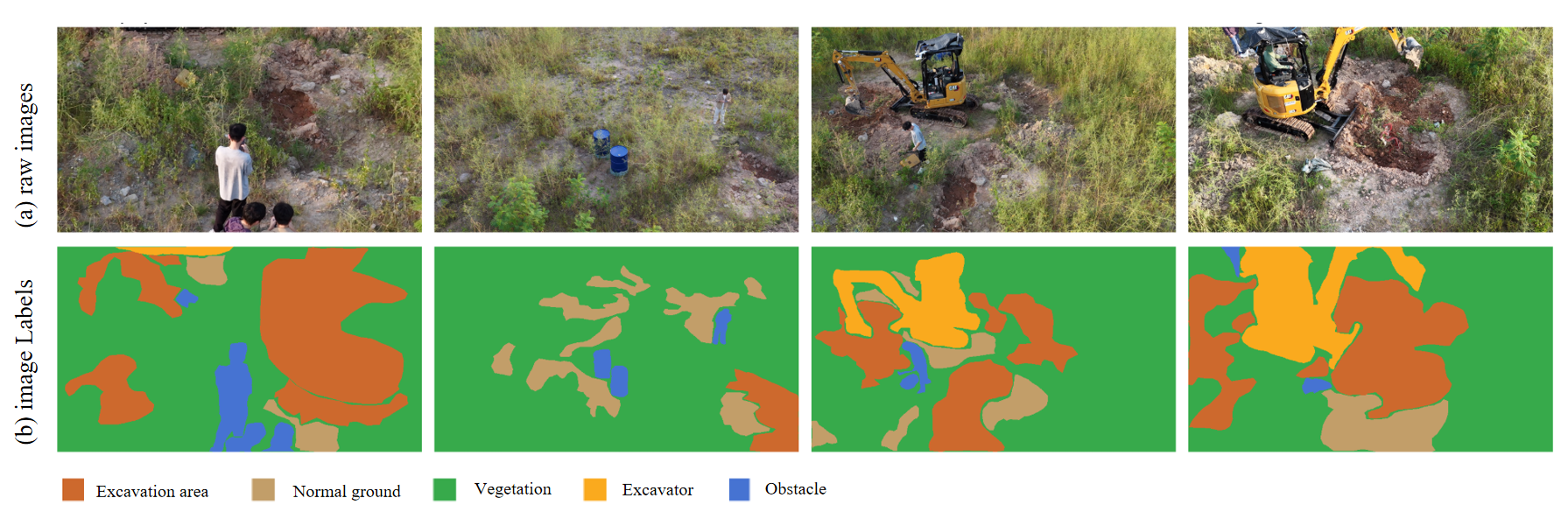}
    \caption{Outdoor Construction Excavation Site (OCES) Dataset}
    \label{fig:oces_dataset}
\end{figure}

\textbf{Outdoor Construction Excavation Site (OCES) Dataset:}
We additionally collect a dataset from real outdoor excavation scenes, comprising 25 video sequences, 5 object categories, and approximately 15k frames. The dataset provides metric object dimensions, LiDAR-based elevation references, and manually annotated semantic segmentation images. Fig.~\ref{fig:oces_dataset} shows a representative example of the collected excavation scenes.

\begin{table*}[h]
\centering
\caption{Comparison of the four evaluation datasets and their roles in this study.}
\label{tab:dataset_comparison}
\small
\renewcommand{\arraystretch}{1.35}
\setlength{\tabcolsep}{5pt}
\begin{tabularx}{\textwidth}{
    >{\raggedright\arraybackslash}p{1.45cm}
    >{\raggedright\arraybackslash}p{1.15cm}
    >{\raggedright\arraybackslash}p{3.55cm}
    >{\centering\arraybackslash}p{1.45cm}
    >{\raggedright\arraybackslash}p{3.05cm}
    >{\raggedright\arraybackslash}X
}
\toprule
\textbf{Dataset} &
\textbf{Source} &
\textbf{Key scale} &
\textbf{Visual input} &
\textbf{Geometric Reference} &
\textbf{Evaluation role} \\
\midrule

\textbf{RELLIS-3D}~\cite{jiang2020rellis3d} &
Public &
13,556 LiDAR scans; 6,235 RGB images; 20 semantic classes &
RGB video &
LiDAR-based terrain reference &
Terrain elevation \\

\textbf{Objectron}~ &
Public &
14,819 annotated video sequences; 9 categories; 180 videos sampled &
RGB video &
Object-level metric reference &
Object dimension recovery \\

\textbf{GOOSE-Ex}~\cite{hagmanns2025gooseex} &
Public &
Selected outdoor sequences with semantic annotations &
RGB video &
-- &
Semantic mapping \\

\textbf{OCES} &
Self-collected &
25 video sequences; 5 object categories; $\sim$15k RGB frames; 3,500 LiDAR scans &
RGB video &
LiDAR-based terrain reference + object dimensions &
\textbf{Object dimension recovery; terrain elevation; semantic mapping} \\

\bottomrule
\end{tabularx}
\end{table*}

\vspace{-0.05cm}
\paragraph{Metrics}
We evaluate geometry using object dimension error and terrain elevation MAE/RMSE, semantics using mIoU and spatial coverage, and excavation performance using the final terrain elevation and excavation volume errors.
\paragraph{Implementation}
Experiments were run on an NVIDIA GeForce RTX 5070 Ti (16~GB), an Intel Core
Ultra 7 265KF, and 15~GiB of visible system memory.

\subsection{Experimental Results}
\paragraph{Quantitative Results}
Tab.~\ref{tab:geometry_accuracy} evaluates the geometric reliability of the
proposed scene representation from two aspects: object-scale recovery and
terrain elevation estimation. On Objectron, the mean absolute error (MAE),
root mean square error (RMSE), and relative error of object dimensions are
2.63~cm, 4.55~cm, and 9.01\%, respectively. On the OCES dataset, the
corresponding errors are 6.23~cm, 8.94~cm, and 10.43\%. For terrain elevation,
the MAE and RMSE are 28.30~cm and 32.14~cm on the selected RELLIS-3D scene.
On the OCES dataset, the corresponding values are 12.61~cm and 15.32~cm,
respectively. These results demonstrate that the framework provides
metric-scale geometric estimates across both object-level and terrain-level
evaluation settings. The reconstruction accuracy is affected by visual
conditions, and the selected RELLIS-3D scene contains lower-quality imagery
and more environmental interference, which introduce additional
reconstruction drift.

\begin{table}[h]
\centering
\caption{Geometric accuracy of the proposed scene representation.}
\label{tab:geometry_accuracy}

\resizebox{\columnwidth}{!}{
\begin{tabular}{@{}llccc@{}}
\toprule
\textbf{Task} & \textbf{Dataset}
& \textbf{MAE (cm)} $\downarrow$
& \textbf{RMSE (cm)} $\downarrow$
& \textbf{Rel. Error (\%)} $\downarrow$ \\
\midrule

\multirow{2}{*}{Object Dimension}
& Objectron & 2.63 & 4.55 & 9.01 \\
& OCES & 6.23 & 8.94 & 10.43 \\

\midrule

\multirow{2}{*}{Terrain Elevation}
& RELLIS-3D & 28.30 & 32.14 & -- \\
& OCES & 12.61 & 15.32 & -- \\

\bottomrule
\end{tabular}
}

\end{table}

\begin{figure}[h]
    \centering
    \includegraphics[width=\columnwidth]{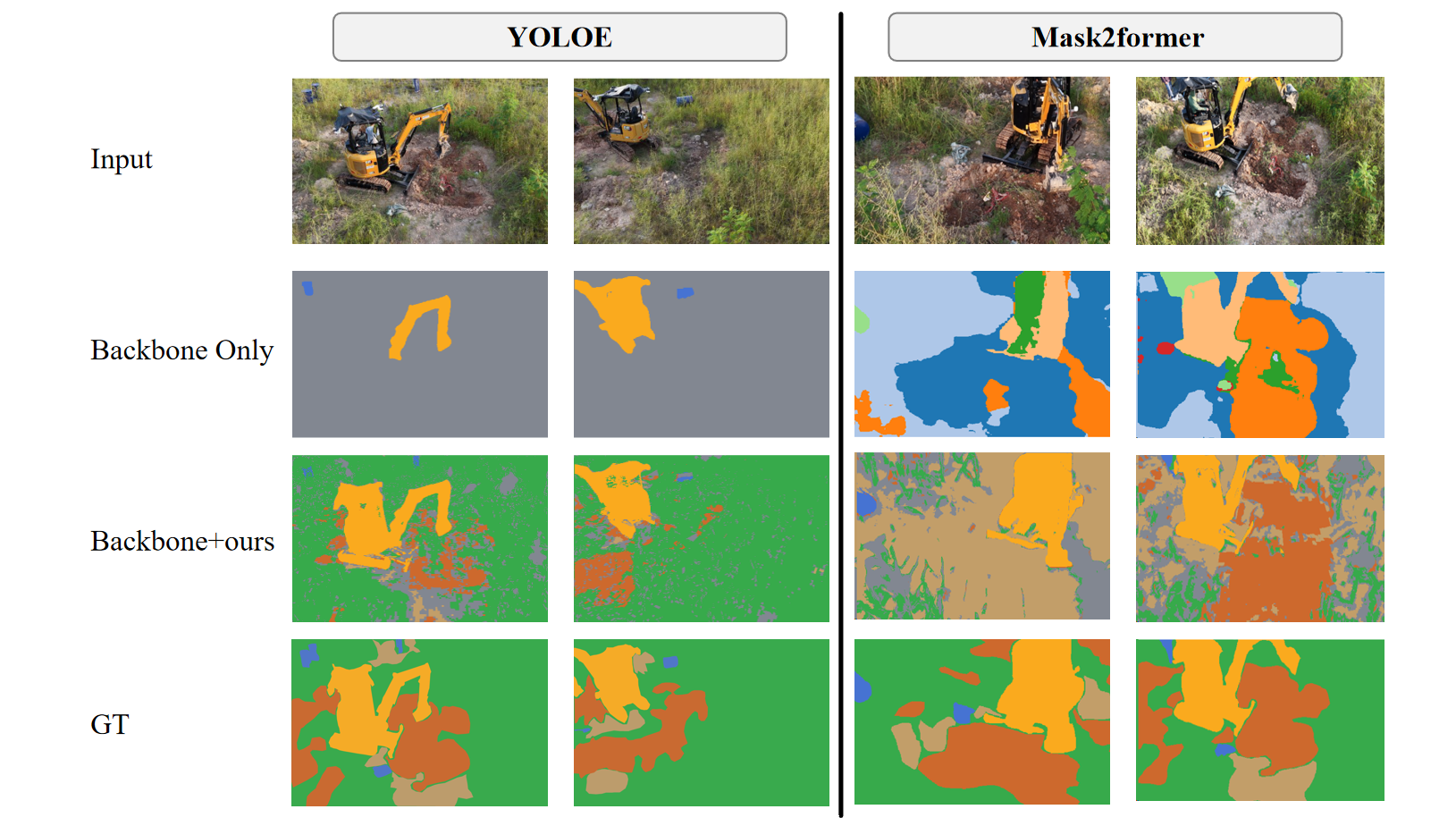}
    \caption{The qualitative comparison of GT, YOLOE, Mask2Former and our method.}
    \label{fig:qualitative}
\end{figure}

We evaluate YOLOE~\cite{wang2025yoloe} and
Mask2Former~\cite{mask2former} as two semantic backbones in our framework.
Without additional task-specific training or fine-tuning, the two models exhibit different characteristics in challenging outdoor scenes. As shown in Tab.~\ref{tab:semantic_accuracy}, YOLOE achieves only
3.34\% mIoU and 2.57\% coverage on GOOSE-Ex when used as the backbone alone,
while the proposed geometry--semantic fusion increases these values to
20.77\% and 90.52\%, respectively. On OCES, YOLOE improves from 6.53\% mIoU
and 2.70\% coverage to 34.57\% and 80.40\%. Mask2Former provides stronger
backbone performance, with 21.40\% mIoU and 94.26\% coverage on GOOSE-Ex and
34.71\% mIoU and 91.85\% coverage on OCES. After geometry-guided fusion,
these values increase to 38.03\% and 96.15\% on GOOSE-Ex and 53.19\% and
99.28\% on OCES. Thus, the proposed method improves both semantic accuracy
and spatial coverage for both backbones, while the magnitude of improvement
depends on the type and severity of the backbone's initial errors. Fig.~\ref{fig:qualitative}
provides a qualitative comparison of the ground truth, backbone predictions,
and the proposed geometry-guided results.

Our framework improves both backbones by exploiting the reconstructed geometry as a common spatial support for semantic refinement. For YOLOE, the main benefit is the recovery of missing regions through multi-view association and spatial propagation, which increases coverage while maintaining consistent object boundaries. For Mask2Former, the improvement mainly comes from geometry-guided filtering and consistency constraints, which suppress isolated or implausible predictions and reduce semantic instability across views. As a result, both semantic accuracy and spatial coverage are improved, demonstrating that the proposed geometry-semantic coupling can compensate for different types of errors produced by generic semantic backbones without requiring additional task-specific training.

\begin{table}[h]
\centering
\caption{Semantic accuracy and spatial coverage.}
\label{tab:semantic_accuracy}


\resizebox{\columnwidth}{!}{
\begin{tabular}{@{}llcccc@{}}
\toprule
& & \multicolumn{2}{c}{\textbf{GOOSE-Ex}}
& \multicolumn{2}{c}{\textbf{OCES}} \\
\cmidrule(lr){3-4}\cmidrule(lr){5-6}

\textbf{Backbone} & \textbf{Method}
& \textbf{mIoU (\%)} $\uparrow$
& \textbf{Coverage (\%)} $\uparrow$
& \textbf{mIoU (\%)} $\uparrow$
& \textbf{Coverage (\%)} $\uparrow$ \\
\midrule

YOLOE & Backbone
& 3.34 & 2.57 & 6.53 & 2.70 \\[2pt]

YOLOE & + Ours
& \textbf{20.77} & \textbf{90.52}
& \textbf{34.57} & \textbf{80.40} \\[2pt]

\midrule

Mask2Former & Backbone
& 21.40 & 94.26 & 34.71 & 91.85 \\[2pt]

Mask2Former & + Ours
& \textbf{38.03} & \textbf{96.15}
& \textbf{53.19} & \textbf{99.28} \\

\bottomrule
\end{tabular}
}

\end{table}

\paragraph{Real-world experiments}

\begin{figure}[h]
    \centering
    \includegraphics[width=\columnwidth]{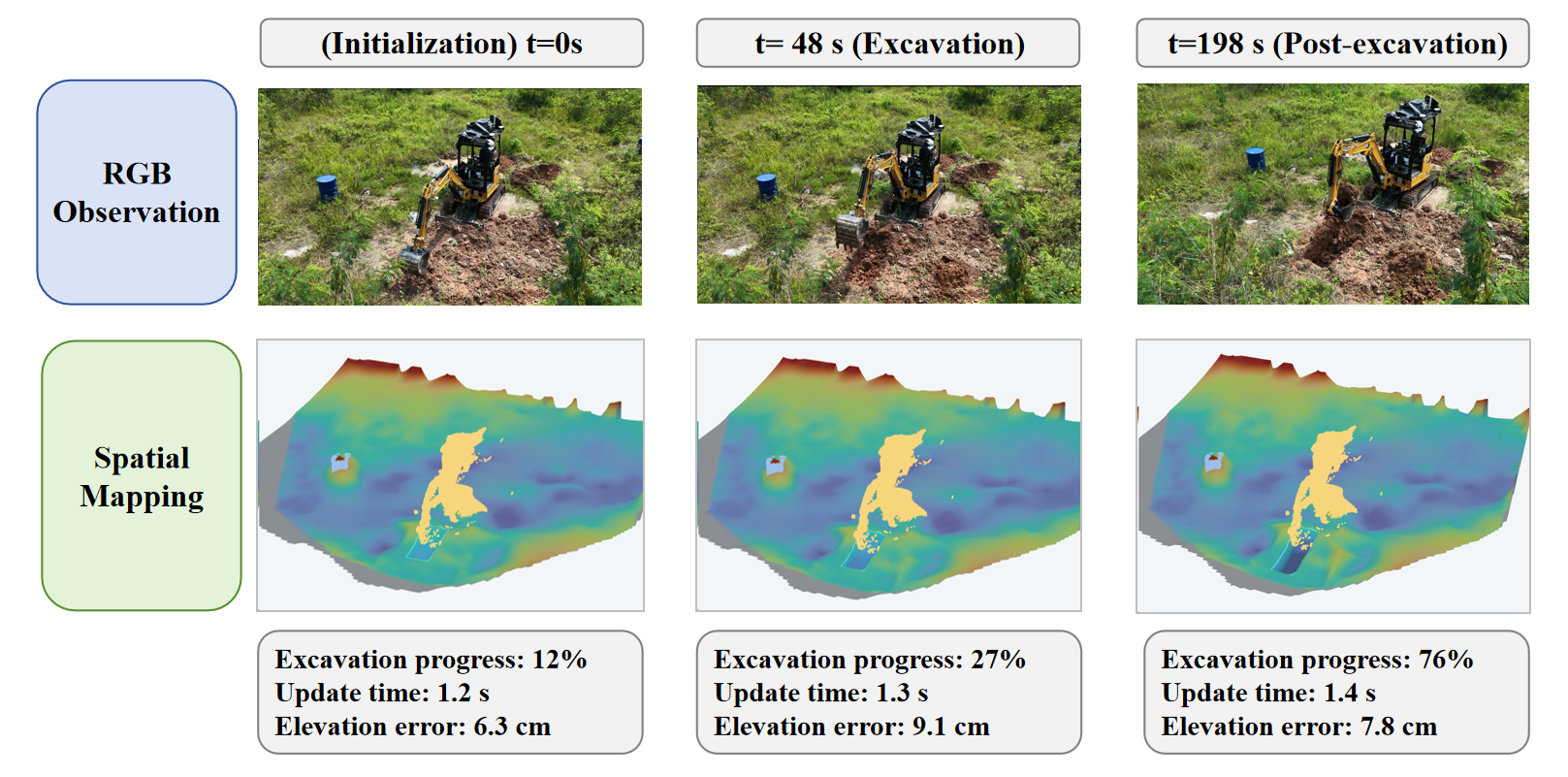}
    \caption{Evolution of the proposed spatial representation during real-world excavation}
    \label{fig:progress}
    \vspace{-4mm}
\end{figure}

To evaluate the practical applicability of the proposed perception system, we integrated it with a Cat 303 CR compact excavator in a real-world excavation experiment. A UAV was positioned at an oblique overhead viewpoint to provide RGB observations of the workspace as the visual input. The excavator was retrofitted with a remote-control interface, allowing motion and digging commands generated by an external planning module to be executed on the real machine. During operation, our system continuously provided task-oriented spatial information, including terrain geometry, semantic regions, and candidate excavation areas, to support the planning and execution of successive digging actions. Fig.~\ref{fig:progress} presents representative snapshots of the excavation process together with the corresponding perception results, illustrating how the spatial representation evolves as the terrain changes during operation.

The system achieved an average spatial-map update interval of approximately 1.4~s, which was sufficient to match the action frequency of the excavator during the tested operation. We further compared the reconstructed elevation map with a LiDAR-based reference throughout the dynamic excavation process. Over regions affected by excavation, the mean elevation error was $12.74 \pm 5.32$~cm, while the minimum short-term mean error reached approximately 5~cm during relatively stable periods. Larger deviations mainly occurred during rapid terrain changes and transient visual disturbances caused by machine motion. 

These results show that the proposed system provides an update rate suitable for practical excavator operation. Using only visual input, it achieves reliable spatial perception at lower sensing cost while jointly providing geometric and semantic information. The resulting representation suppresses environmental interference and preserves task-relevant structure, supporting outdoor excavation tasks such as site clearing, navigation, and digging.

\vspace{0.05cm}
\vspace{0.05cm}

\subsection{Ablation study}
The ablation results highlight the importance of local geometric filtering and semantic spatial consistency processing. Removing local geometric filtering reduces mIoU by 11.65\% and 12.75\% on GOOSE-Ex and OCES. Removing semantic spatial consistency processing causes the largest coverage reductions, reaching 27.34\% on GOOSE-Ex and 29.93\% on OCES, while also decreasing mIoU by 10.41\% and 16.65\%. These results suggest that geometric filtering and spatial consistency play complementary roles in improving semantic accuracy and maintaining spatial coverage.
\begin{table}[h]
\centering
\caption{Ablation study.}
\label{tab:yoloe_ablation}

\setlength{\tabcolsep}{3pt}
\renewcommand{\arraystretch}{1.05}

\resizebox{\columnwidth}{!}{
\begin{tabular}{@{}lcccc@{}}
\toprule
& \multicolumn{2}{c}{GOOSE-Ex}
& \multicolumn{2}{c}{OCES} \\
\cmidrule(lr){2-3}\cmidrule(lr){4-5}
\textbf{Configuration}
& \textbf{mIoU}
& \textbf{Cov.}
& \textbf{mIoU}
& \textbf{Cov.} \\
\midrule

YOLOE backbone
& 3.34 & 2.57 & 6.53 & 2.70 \\[2pt]

\midrule

w/o Local geometry
& 9.12 & 70.46 & 21.82 & 64.13 \\[2pt]

w/o Terrain continuity
& 17.84 & 81.28 & 29.63 & 79.72 \\[2pt]

w/o Object support
& 18.40 & 89.07 & 30.28 & 81.94 \\[2pt]

w/o Spatial consistency
& 10.36 & 63.18 & 17.92 & 50.47 \\[2pt]

w/o View reliability
& 19.11 & 90.87 & 32.18 & 78.06 \\[2pt]

\midrule

YOLOE + ExcavaTwin (full)
& 20.77 & 90.52 & 34.57 & 80.40 \\

\bottomrule
\end{tabular}
}

\vspace{2pt}
{\footnotesize
Cov. denotes coverage; all values are percentages.
}

\end{table}

\section{CONCLUSIONS}
\textbf{Conclusion.}
We propose ExcavaTwin, a novel perception framework for autonomous excavators working in outdoor environments. The proposed framework extracts geometric structure from a visual reconstruction model and semantic masks from a segmentation model, coupling the two to suppress outdoor interference and build a comprehensive and reliable scene representation.The resulting spatial map jointly maintains geometric and semantic information while suppressing irrelevant environmental disturbances.

Experiments across four datasets and a real excavation platform demonstrate
the reliability and practical applicability of the proposed representation.
The framework obtains object-dimension MAE/RMSE of 2.63/4.55~cm on Objectron
and 6.23/8.94~cm on OCES, while terrain-elevation MAE/RMSE reaches
12.61/15.32~cm on OCES. For semantic mapping, the proposed fusion increases
mIoU to 38.03\% on GOOSE-Ex and 53.19\% on OCES with Mask2Former, while the
real-world system maintains an average update interval of approximately
1.4~s and a dynamic-region mean elevation error of
$12.74 \pm 5.32$~cm.

\textbf{Limitations.}
The current study only demonstrates the potential of the proposed representation to support downstream tasks such as site cleanup and navigation, while its effectiveness in these tasks has not yet been validated on a real excavator. The effects of UAV viewpoint and camera-to-scene distance on reconstruction accuracy remain to be systematically investigated.

\textbf{Future work.}We believe that vision-based perception is a promising direction for autonomous excavation. Future work will explore integrating models such as VLMs to achieve tighter coupling between visual perception, planning, and control, thereby improving the autonomy and intelligence of excavation systems.




\section*{ACKNOWLEDGMENT}

The authors used ChatGPT to assist with language editing and polishing of the manuscript. All revisions were reviewed and approved by the authors, who take full responsibility for the final content.

\end{document}